\documentclass[conference]{IEEEtran}

\usepackage{amsmath,amsfonts}
\usepackage{algorithmic}
\usepackage[T1]{fontenc}  
\usepackage[utf8]{inputenc}  

\usepackage{tabularx}
\usepackage{array}
\usepackage{ragged2e}

\usepackage{booktabs}
\usepackage{algorithm}
\usepackage{algorithmic}
\usepackage{array}
\usepackage{xcolor}
\usepackage{amssymb}
\usepackage[caption=false,font=normalsize,labelfont=sf,textfont=sf]{subfig}
\usepackage{textcomp}
\usepackage{multirow}
\usepackage{stfloats}
\usepackage{url}
\usepackage{verbatim}
\usepackage{graphicx}
\usepackage{cite}
\begin{document}

\title{Position Paper: Neurotransmitters as a Missing Dimension in Artificial Neural Networks}
\author
{
\hspace*{-1cm}\IEEEauthorblockN{1\textsuperscript{st} Yupei Li}
\IEEEauthorblockA{\hspace*{-1cm}\textit{Department of Computing and Chair of Health Informatics} \\
\hspace*{-1cm}\textit{Imperial College London and Technical University of Munich}\\
\hspace*{-1cm}London and Munich, United Kingdom and Germany \\
\hspace*{-1cm}yl7622@ic.ac.uk}
\and



\hspace{-2cm}\IEEEauthorblockN{2\textsuperscript{nd} Manuel Milling}\hspace{8cm}
\IEEEauthorblockA{\hspace{-2cm}\textit{Chair of Health Informatics} \\
\textit{\hspace{-2cm}Technical University of Munich}\\
\hspace{-2cm}Munich, Germany \\
\hspace{-2cm}manuel.milling@tum.de}
\and
\IEEEauthorblockN{3\textsuperscript{rd} Berrak Sisman}
\IEEEauthorblockA{\textit{Department of Electrical and Computer Engineering} \\
\textit{Johns Hopkins Whiting School of Engineering}\\
Baltimore, United States \\
sisman@jhu.edu}
\and

\IEEEauthorblockN{4\textsuperscript{th} Bj\"orn Schuller}
\IEEEauthorblockA{\textit{Chair of Health Informatics and Department of Computing} \\
\textit{Technical University of Munich and Imperial College London}\\
Munich and London, Germany and United Kingdom \\
schuller@tum.de}
}

\markboth{Journal of \LaTeX\ Class Files}%
{Shell \MakeLowercase{\textit{et al.}}: Neurotransmitters as a Missing Dimension in Artificial Neural Networks}

\IEEEpubid{
  \parbox{\textwidth}{ 
    \centering 
    ~\copyright~2026 The Authors. This work is licensed under a Creative Commons Attribution-NonCommercial-NoDerivatives 4.0 License.\\
    For more information, see \url{https://creativecommons.org/licenses/by-nc-nd/4.0/}
  }
}


\maketitle
\begin{abstract}
Artificial neural networks (ANNs), as core components of modern deep learning (DL) systems, lack the adaptive flexibility and long-term stability exhibited by biological systems. This limitation largely stems from the fact that conventional ANNs rely on uniform, local, and gradient-based parameter updates, while neglecting internal learning principles that are biological mechanisms such as neurotransmitters signalling or neuroplasticity. Consequently, many existing approaches focus on architectural expansion or mathematical fine-tuning techniques such as regularisation or parameter isolation. Inspired by the superior adaptability and plasticity of mammalian brains, we posit that neuromodulation with neurotransmitters constitutes a third axis of learning, complementary to neural activity and synaptic plasticity, and should be explicitly modelled in artificial neural networks. In this positional paper, we argue that incorporating neuromodulatory principles into ANN design represents a promising and underexplored research direction, and we advocate for greater attention to this perspective in the development of adaptive and continual learning systems.
\end{abstract}

\begin{IEEEkeywords}
Neural Networks, Neurotransmitters, Activation, Adaptive learning, Continuous learning
\end{IEEEkeywords}

\section{Introduction}
Recent advances in deep learning (DL) are predominantly based on artificial neural networks (ANNs), whose strong empirical performance particularly in large language models (LLMs) provides compelling evidence of their effectiveness. These performance gains are often attributed to the Universal Approximation Theorem \cite{cybenko1989approximation}, which states that NNs can approximate a broad class of functions and thereby model complex real-world data distributions. This theoretical foundation has driven a major research trend that builds upon mathematical principles, resulting in models that exhibit high-level capabilities such as reasoning and language understanding.

The biological origins of neural networks, namely their inspiration from the mammalian brain biological mechanisms, have gradually been de-emphasised, such as neurotransmitter signalling and neuroplasticity, and the learning processes used to endow NNs with problem-solving abilities have become increasingly simplified. Contemporary training paradigms rely heavily on gradient-based optimisation with relatively rigid optimisers \cite{nwankpa2020advances}, which can lead to training instability and, in multitask or continual learning settings, catastrophic forgetting \cite{ashley2021doesadamoptimizerexacerbate}. Although adaptive techniques, such as adaptive learning strategies (e.g., adjusting the learning rate \cite{iiduka2021appropriate, jepkoech2021effect}), have been proposed to mitigate issues arising from overly simplified approaches to sequential learning, these methods tend to address only the symptoms rather than the underlying challenges \cite{martin2020systematic}.

The challenges are that the majority of current methods are still principly based on a fundamentally loss-centric understanding of learning, which is defined as ``minimise a scalar loss via gradient descent'' \cite{terven2025comprehensive, gambella2021optimization}. With all parameters subject to the same update rules and only varying in magnitude, learning in this paradigm is essentially reduced to the uniform minimisation of a scalar objective, even under consideration of elaborate regularisation techniques. Although these formulations have worked well in static or well-defined environments, they implicitly assume that learning should proceed consistently and continuously in the presence of an error signal. What is required is a form of plasticity that governs not only what is learnt, but also whether, where, and to what extent learning should occur, in a manner that is tightly coupled to both internal states and environmental context \cite{novak1984learning}. This requires moving beyond optimisation alone and rethinking learning as a regulated, context-sensitive process rather than a purely objective-driven one.

In contrast, the mammalian brain employs a far more sophisticated learning process, selectively consolidating information into long-term memory while adaptively forgetting less relevant content to maintain efficiency, a phenomenon well supported by synaptic plasticity theory \cite{fuchs2014adult}. Re-examining and incorporating these biologically inspired learning principles may therefore offer a promising direction.

\IEEEpubidadjcol
One aspect of mammalian brain may inspire efforts to move from rigid neural network structures towards dynamic, adaptive processes is neuroplasticity 
\cite{li2025neuroplasticityartificialintelligence}. 
However, existing ANN formulations of neuroplasticity typically emphasize uniform or global learning mechanisms, resulting in learning strategies that remain largely one-size-fits-all across parameters and tasks. This raises an important concern, as it risks concentrating research too narrowly on representation, instead of training strategy such as adaptative learning (plasticity) or modulation. 

In contrast, in mammalian brains, when and how learning is performed is, to a large part, controlled through
neurotransmitters, which are key molecules responsible for neuromodulation and are essential for information transmission. More precisely, a \textbf{neurotransmitter is a chemical signalling molecule secreted by a neuron that crosses a synapse to transmit a signal to another cell, which may be another neuron, a muscle cell, or a gland cell} \cite{axelrod1974neurotransmitters}.
They play critical roles in diverse physiological and behavioural functions, including emotion regulation, and dysfunction in neuromodulation can lead to severe mental disorders such as depression \cite{bryan2023chemical}. By providing a mechanism for flexible and stable regulation of short- and long-term memory, neurotransmitters suggest a potential third axis of learning in artificial neural networks, complementary to neural activity and synaptic plasticity.

Therefore, in this positional paper, we outline the current understanding of neurotransmitter-driven biological learning and propose a conceptual framework for incorporating neuromodulatory principles into ANNs.

\section{Neurotransmitters in Biological Learning}
\label{sec:2}
\begin{figure}[t]
    \centering
    \includegraphics[width=0.85\linewidth]{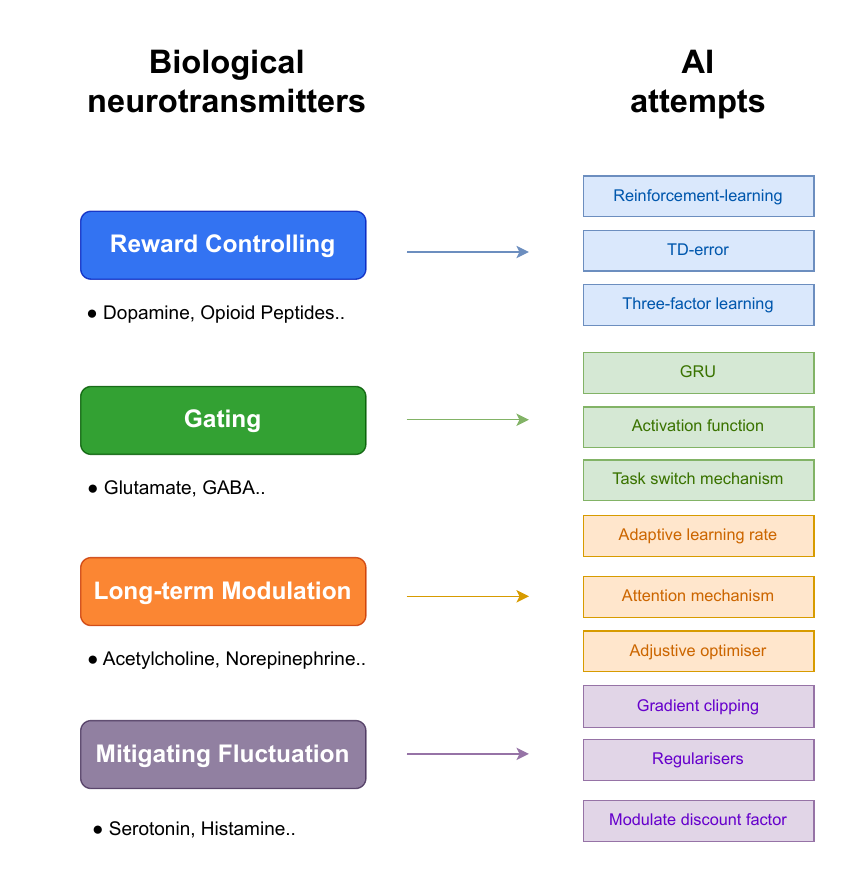}
    \caption{Conceptual illustration connecting biological neurotransmitter types, associated cases, and corresponding AI approaches.}
    \label{fig:conceptual}
\end{figure}

The functionality of neurotransmitters reaches beyond the previously mentioned chemical modulation of neural learning including a range of roles, shown in a conceptual figure \ref{fig:conceptual}.

First and foremost, neurotransmitters are essential to  \textbf{reward controlling (RC)}. 
For instance, dopamine is a key neurotransmitter famously associated with positive affect and reward processing in humans. It is released by a midbrain neuromodulatory system that projects broadly to subcortical and cortical regions. Rather than conveying detailed sensory or motor information, dopamine provides a global teaching signal that modulates synaptic plasticity through some types of receptors. In motor circuits, this modulation regulates action selection and execution, while in limbic and cortical circuits it supports reward-based learning by reinforcing actions and states that lead to better-than-expected outcomes \cite{schultz2016dopamine}. Dopamine, a neurotransmitter, could shape the mammalian behaviour from immediate signals between neurons on a micro-level. For example, experiments in rats have demonstrated that selective optogenetic activation of midbrain neurons affected by dopamine, even in the absence of actual rewards, is sufficient to assign reward-motivational value to originally neutral conditioned stimuli. This phasic discharge of dopamine neurons consequently drives rodents to exhibit approach behaviour \cite{Saunders2018Dopamine}. These findings suggest that dopamine signalling provides a natural biological inspiration for computational mechanisms of reward modulation, particularly in reinforcement learning frameworks within artificial neural networks, as discussed in Section \ref{sec:relation}.

\textbf{``Gating (G)''} represents another fundamental role of neurotransmitters in a neural network, with GABA serving as a representative example. It can effectively be interpreted as the controller for a gating mechanism~\cite{Farrant2005}. It is an inhibitory neurotransmitter that temporarily suppresses neuronal activity, reducing excitability and preventing overactivation. Neurons can become active again once the inhibition subsides. Gating mechanisms have also been widely adopted in ANNs, most notably in architectures such as long short-term memory (LSTM) networks \cite{hochreiter1997long}. However, in current ANN implementations, gating thresholds are determined solely through automatic parameter optimisation, without an explicit modulatory stimulus analogous to neurotransmitter signalling. By contrast, gating in the brain is dynamically regulated through neuromodulatory systems operating across multiple temporal scales, enabling flexible, state-dependent reconfiguration of information flow that cannot be reduced to a single learnt gating function, as in LSTM architectures. This gap highlights a potential opportunity for incorporating neurotransmitter-inspired modulation as an explicit control signal for gating dynamics in ANNs.

Beyond binary gating mechanisms, neurotransmitters also support \textbf{long-term, continuous modulation (LM)}, it is released by a small group of neurons that broadcast a global modulatory signal to many brain regions. Functionally, acetylcholine acts as a neuromodulatory signal that modulates cortical state and information routing. It is released primarily by cholinergic neurons in the basal forebrain and acts on nicotinic and muscarinic receptors in the cortex and hippocampus to modulate attention, sensory processing, and cognitive flexibility. Proper acetylcholine signalling facilitates adaptive information flow and learning by enhancing stimulus-driven representations while suppressing interference from recurrent or irrelevant activity. Conversely, disruptions in cholinergic modulation have been associated with impairments in attentional control, learning, and flexible cognitive processing \cite{hasselmo2006role}. Norepinephrine is another example that regulates how strongly neural signals are amplified relative to background noise. Appropriate levels of it increase alertness and attention by improving the signal-to-noise ratio of neural responses, thereby making the system more sensitive to salient or unexpected inputs \cite{astonjones2005integrative}. Analogous strategies are already implicitly employed in ANNs through optimisation techniques such as cosine adaptive learning rate schedules \cite{verma2025sine} and decoupled weight decay in AdamW \cite{loshchilov2019decoupled}. However, these mechanisms rely on predetermined algorithmic schedules that specify when and how strongly adjustments are applied, offering only a coarse approximation of biological neuromodulation. This contrast suggests that more heuristic, state-dependent approaches inspired by norepinephrine-mediated modulation may warrant further investigation.

When changes in certain states or network elements occur too rapidly, other neurotransmitters can act to \textbf{mitigate these fluctuations (MF)}. Serotonin provides a representative example. It provides a complementary form of neuromodulation. Rather than encoding specific information, it exerts a global regulatory influence through multiple receptor types. Functionally, serotonin dampens excessive neural responses, reduces sensitivity to short-term fluctuations, and promotes long-term behavioural stability. From a dynamical systems perspective, it constrains the system’s effective degrees of freedom, thereby preventing impulsive or unstable dynamics \cite{dayan2008serotonin}. In ANNs, analogous effects are commonly achieved through stabilisation techniques, such as gradient clipping \cite{chen2020understanding}, which act as moderation to prevent abrupt changes on gradient that may lead to sudden change of weight to crush the whole training. However, the underlying principles differ: whereas serotonin maintains stability in response to environmental fluctuations, standard ANN gradient clipping operates solely during training as an external optimisation constraint. This difference highlights a potential avenue for further investigation into more adaptive, biologically inspired stabilisation mechanisms in artificial networks.

There exists a vast array of neurotransmitters, many of which fall within the functional scope discussed above. For the sake of an overview, Table \ref{tab:pragmatic mapping} summarises important ones along with their function in the mammalian brain, inspiration of application in the artificial neural network context, and references to AI attempts to incorporate these.

Importantly, neurotransmitters do not carry explicit informational content, but they serve as modulatory elements that influence other neural representations. Moreover, they can interact synergistically to maintain a balanced biological system \cite{raiteri2002coexistence}. This observation provides a potential inspiration for ANNs: instead of relying solely on a single loss signal, considering the overall dynamics of the learning process may promote more stable and robust learning. Furthermore, neuromodulatory mechanisms could contribute to effective memory management in the brain through selective forgetting \cite{Berry2020DopamineForgetting}. Drawing inspiration from these principles could potentially mitigate some of the severe challenges faced by current deep learning systems.

\begin{table*}[t]
\centering
\small
\setlength{\tabcolsep}{4pt}
\renewcommand{\arraystretch}{1.25}
\begin{tabularx}{\textwidth}{
  >{\RaggedRight\arraybackslash}p{2.8cm}
  >{\RaggedRight\arraybackslash}X
  >{\RaggedRight\arraybackslash}X
  >{\RaggedRight\arraybackslash}p{3.5cm}
}
\toprule
\textbf{Neurotransmitter / neuromodulator} &
\textbf{High-level function in brains} &
\textbf{How to imitate in deep nets (AI-native mapping)} &
\textbf{AI attempts / references} \\
\midrule

Dopamine and Opioid Peptides or Substance P &
(RC) reward prediction error; reinforcement learning; pain relief or salience; &
TD-error signals; three-factor learning (pre, post and modulation learning); neuromodulated plasticity to adjust where and when learning happens &
Three-factor learning overview \cite{Kusmierz2017ThreeFactor}; TD-reinforcement learning \cite{li2019reinforcement}. \\

Oxytocin and Vasopressin &
(RC) social trust or dominance; prosocial bias in social contexts &
Multi-agent RL, cooperation or competitive priors, reward shaping for prosocial outcomes &
Trust-based consensus multi-agent system \cite{fung2022trust}. \\

Glutamate &
(G) primary fast excitation; drives most forward and feedback signal; supports associative plasticity (e.g., NMDA-dependent) &
Standard excitatory weights or activations; coincidence-style plasticity via gated Hebbian terms;  &
Hebbian learning with gates \cite{aubin1998hebbian}. \\

GABA and Glycine &
(G) primary fast inhibition; stabilises dynamics; gain control; prevents runaway excitation &
Explicit inhibitory constraints (Dale-like sign structure), lateral inhibition, competitive dynamics & Dale-compliant ANNs \cite{Cornford2021Dale}; gated-plasticity framework \cite{Kusmierz2017ThreeFactor}. \\

Epinephrine&
(G) energy mobilisation; wakefulness; arousal regime switch &
Task-level mode switch; brief large learning-rate bursts &
One time learning through strengthening entire patterns of activation \cite{remmelzwaal2020biologicallyinspiredsalienceaffectedartificial}. \\

Acetylcholine and Norepinephrine &
(LM) attention, arousal learning; flexible cognitive processing; adaptive gain; linked to unexpected uncertainty &
Attention-temperature modulation, context-dependent modulation; &
Uncertainty/neuromodulation theory \cite{YuDayan2005Uncertainty}; Attention mechanism \cite{attention}. \\

Endocannabinoids and Somatostatin &
(LM) retrograde signaling; decreases presynaptic release; regulates plasticity; controls excitability &
Local retrograde suppression of updates; anti-runaway stabilisation; &
Signals collaborating or modulating with other signals \cite{mei2025improvingadaptivecontinuouslearning} \\

Nitric Oxide (NO) &
(LM) diffusive modulation; spatially broad influence on plasticity &
Diffusive modulatory fields that locally gate learning rates; promote modularity; &
Diffusion-based neuromodulation to reduce catastrophic forgetting \cite{Velez2017DiffusionNeuromod}. \\

Adenosine Triphosphate (ATP)  &
(LM) energy-state signaling; not directly joining for modulation &
Energy budget variables; adaptive compute (early exit), penalties on updates &
Neuromodulation‑inspired learning rate modulation \cite{razmi2022adaptive}. \\

Serotonin (5-HT) &
(MF) patience or impulsivity; behavioural inhibition; &
Modulate discount factor, decision thresholds; stabilise updates by reducing volatility &
Learn online via meta-gradients \cite{Xu2018MetaGradientRL}. \\

Histamine, Adenosine and Neuropeptide Y &
(MF) sleep status modulation; homeostatic protection; stress resilience; &
Activity-dependent fatigue (penalise high activation), online and offline consolidation schedules, activation-cost regularisers; stabilise policies; reduce volatility &
Synaptic related adjustive methods \cite{HofmannMader2022SynapticScaling}. \\

\bottomrule
\end{tabularx}
\vspace{0.1cm}
\caption{Pragmatic mapping from major neurotransmitters/neuromodulators to deep learning design patterns. RC, G, LM, and MF denote four functional roles summarised in Section \ref{sec:2}: reward control, gating, long-term modulation, and fluctuation mitigation, respectively.The ``AI attempts'' column cites representative prior work where closely related mechanisms have been implemented or formalised.}
\label{tab:pragmatic mapping}
\end{table*}

\textbf{A useful unifying pattern (implementable design).}
A clean abstraction that covers most rows above is a \emph{two-network} or \emph{two-stream} design:
\begin{itemize}
  \item \textbf{Base network} (predictor / policy / value model) with parameters $\theta$.
  \item \textbf{Modulator network} producing time- and/or context-dependent signals (e.g., per-layer gain $g_\ell(t)$, attention temperature $\tau(t)$, learning-rate multipliers $\eta_\ell(t)$, or plasticity gates $m_{ij}(t)$).
\end{itemize}
Plasticity is then expressed as a \emph{three-factor} update:
\[
\Delta w_{ij}(t) \propto \underbrace{f(\text{pre}_i(t), \text{post}_j(t))}_{\text{eligibility / Hebbian trace}} \cdot
\underbrace{M(t)}_{\text{global or regional modulator}},
\]
where $M(t)$ can be learnt (backpropagated) as in differentiable neuromodulated plasticity \cite{Miconi2019Backpropamine}, or hand-designed (e.g., Temporal Difference (TD)-error, uncertainty, gain) following established neuromodulation theories \cite{YuDayan2005Uncertainty,AstonJonesCohen2005AdaptiveGain,Kusmierz2017ThreeFactor}, and eligibility trace $f(\cdot)$ depends on the interaction between the presynaptic activity $\text{pre}_i(t)$ from the presynaptic neuron $i$ and the postsynaptic activity $\text{post}_j(t)$ from the postsynaptic neuron $j$. The $w_{ij}(t)$ is the weight connecting the two neurons.

\section{Neurotransmitters as Computational Modulators in NNs}

\subsection{Limitations of current NN Learning Paradigms}

As discussed above, current NN learning paradigms largely follow a rigid gradient descent-based optimisation principle centred on a single loss value. However, many contemporary applications require lifelong learning capabilities, which are severely limited by said approaches. LLMs, for instance, are increasingly deployed as general-purpose tools and are expected to accommodate vast numbers of tasks sequentially. This setting aligns with the objectives of continual or lifelong learning \cite{liu2017lifelong}.

Optimising a single objective across diverse and evolving tasks can lead to conflicting gradient directions, which in turn results in suboptimal learning outcomes \cite{schuch2019monitoring}. Moreover, parameter updates in most NNs are applied uniformly, such that all parameters are adjusted according to the same update rule derived from the final objective, regardless of their functional roles. While adaptive optimisation methods or learning rate schedules can introduce parameter-wise scaling, these mechanisms remain static or pre-specified and do not provide role-specific, state-dependent modulation of parameter updates. This uniformity persists even though different components, such as convolutional layers and linear layers, process information according to distinct computational principles.

Such an approach stands in contrast to biological learning, where plasticity is both region- and mechanism-specific. For example, synaptic plasticity in the hippocampus is predominantly N-methyl-D-aspartate (NMDA)-dependent, whereas plasticity in the striatum requires dopaminergic signalling to initiate learning and memory renewal. Furthermore, forgetting is typically treated as a critical failure mode in artificial neural networks, with numerous strategies proposed to mitigate catastrophic forgetting \cite{rahaman2025dynamic}. In biological systems, however, forgetting is a natural and adaptive process, essential for managing limited memory resources, provided that the system can selectively determine what to retain and what to discard.

\subsection{Proposed methodology}

\begin{figure}
    \centering
    \includegraphics[width=0.9\linewidth]{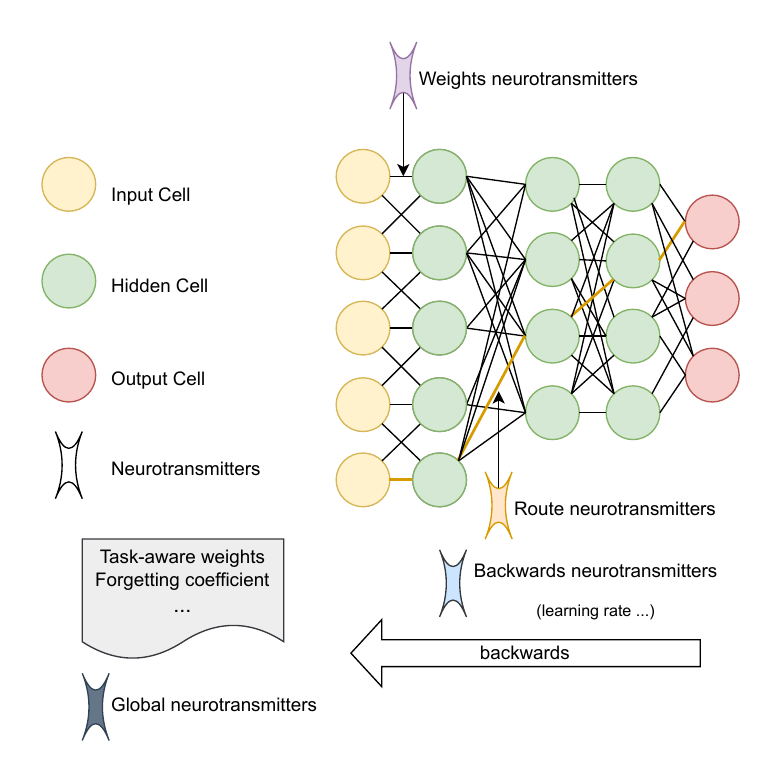}
    \caption{Inspired modulation in a simplified artificial neural network. Multiple classes of artificial ``neurotransmitters'' are illustrated, each associated with a distinct functional role. At a global level, modulatory signals may influence task prioritisation by scaling or reweighting task-related objectives. At a local level, weight-modulating neurotransmitters can regulate the importance and update dynamics of specific parameters during task learning, while routing-related neurotransmitters selectively gate subsets of neurons involved in particular tasks. In addition, backward or learning-related neurotransmitters may dynamically adjust learning rates or update magnitudes during optimisation. Crucially, all neurotransmitter signals are assumed to be released in a state-dependent manner, and to dynamically modulate the learning process to resemble biological neuromodulatory mechanisms.}
    \label{fig:structure}
\end{figure}

Inspired by learning mechanisms in the mammalian brain, we conceptualise \textbf{artificial neurotransmitters as state-dependent modulators of learning, memory, and inference in artificial neural networks}. Similar to their biological counterparts, these signals do not encode task-specific information directly, but instead regulate how learning and computation are carried out. Specifically, neurotransmitter-inspired modulation may dynamically adjust learning rates at both local and global scales, reweigh loss functions in a task-aware manner, regulate parameter plasticity by distinguishing between stable and adaptable weights, and control attention or routing mechanisms to govern information flow. Collectively, these dimensions define a neuromodulatory axis that complements neural activity and synaptic plasticity, enabling more adaptive, stable, and lifelong learning dynamics. These concepts are highlighted in the proposed visualisation in Figure \ref{fig:structure}.

Specifically, neurotransmitter-inspired signals may be modelled either as gating mechanisms or as continuous, pre-defined or learnable parameters. We further categorise these signals into two types: global and local transmitters. Global transmitters operate at the system level, modulating the overall learning dynamics of the network, whereas local transmitters exert influence over specific components or subsystems, selectively regulating parts of the learning process.

For global neurotransmitters, the importance of a task can be represented by a learnable scalar weight that modulates the overall learning objective:
\begin{equation}
    L = \sum_T f(s_T, T)L_T,
\end{equation}
where $L_T$ is the individual task $T$ loss, and the task importance score or forgetting coefficient could be generated by a global neuromodulatory function conditioned on the learning state $s_T$ and the task itself $T$. The learning state represents the current internal state of the network, including parameter values, optimisation status, neuronal activity (e.g., activation patterns) and so on to represent the internal environment.

For local neurotransmitters, modulation may occur at multiple sites, analogous to the diverse loci of neuromodulatory effects in the mammalian brain. We identify three representative forms of local modulation.

\textbf{Weight neurotransmitters} operate at the level of individual parameters, capturing the notion that different weights encode information of varying importance. Conditioned on the current learning state and environmental context, such modulatory signals can adaptively regulate the degree to which each weight is updated, allowing certain parameters to remain stable while others remain highly plastic. The weight update could thus, for instance, follow the behaviour as shown below:

\begin{equation}
    w^{t+1} = f_w(s^t)(w^t - \sigma \Delta_w L),
\end{equation}

where $\sigma$ is the learning rate and $f_w(s^t)$ shows the adjustable weight decay based on weights and states. The superscript $t$ indexes the training iteration. The equation is not indexed by task $T$, as it applies uniformly to all tasks and describes the learning dynamics within an individual task. The same is true for the latter equation representation.

At a larger scale, pathway-level \textbf{route neurotransmitters} can modulate information flow across distinct network components. This is analogous to architectures such as Mixture-of-Experts \cite{shazeer2017outrageously_iclr} or multi-head transformers \cite{attention}, which allocate different components to process different information. However, unlike these models, in which component specialisation emerges implicitly, pathway-level route neurotransmitters explicitly regulate the routing of information by modulating the activity of specific pathways. The formula is shown as below: 

\begin{equation}
w^{t+1}_{ij}
=
w^{t}_{ij}
-
\sigma \, \mathbf{1}_{(i,j)\in p}\, \nabla_{w_{ij}} L ,
\end{equation}

where
\(
\mathbf{1}_{(i,j)\in p} = 
\begin{cases}
1, & \text{if } (i,j)\in p, \\
0, & \text{otherwise}.
\end{cases}
\),  \\
showing that learning updates are restricted to edges along the active path $p$, reflecting selective, pathway-specific plasticity rather than global parameter adaptation. In other words, if $i^{th},j^{th}$ neurons are on the selected path, the value of $f_p(s^t)$ should be higher.

One additional degree of freedom in the preceding formulation is the learning rate itself. Inspired by biological neuromodulation, the learning rate can be dynamically modelled and modulated through \textbf{backward neurotransmitter} signals. While adaptive learning rate methods exist in ANNs, they are typically governed by pre-defined schedules or heuristics rather than being directly controlled by a biologically inspired modulatory signal. Incorporating a neurotransmitter-like mechanism to regulate the learning rate would allow the network to adjust its plasticity in a state-dependent manner, complementing weight-level and pathway-level modulation. The formula is shown as below: 

\begin{equation}
    w^{t+1}_{ij} = w^t_{ij} - \sigma f_l(s^t) \Delta_w L, 
\end{equation}

where $\sigma$ is the learning rate and $f_l(s^t)$ shows the adjustable weights decay based on the state.

In principle, additional neurotransmitter-inspired signals could be incorporated, each represented as either pre-defined or learnable parameters. This framework is intended as a conceptual draft, highlighting a potential research direction for future exploration. By introducing hierarchical, state-dependent neuromodulation, ANNs may achieve more flexible and stable learning dynamics, offering a promising avenue for performance improvement compared with the rigid, single-objective optimisation strategies commonly employed today.

\subsection{Relation to Existing Learning Paradigms}
\label{sec:relation}

As previously hinted at, there are similar techniques that exist in the current DL field, which, however, show fundamental differences to the proposed neurotransmitters.
First, neurotransmitters provide \textbf{state-dependent, adaptive learning paradigms}. Many optimisation algorithms have explored this domain; for example, Adam \cite{kingma2015adam} and RMSProp \cite{hinton2012rmsprop} employ self-adaptive learning rate schedules based on the current gradient squared, which can be conceptualised as a representation of state. In addition, curriculum learning often adopts a self-paced learning strategy to regulate the progression of learning, thereby preventing the learner from encountering overly difficult cases too early, which could impede learning \cite{kumar2010selfpaced}. However, these existing adaptive mechanisms condition learning primarily on optimisation statistics rather than on internal ``cognitive'' or environmental states, thereby limiting their capacity to modulate learning in a context-sensitive manner.

With respect to state interactions, \textbf{reinforcement learning methods} often rely on rewards derived from the environment, such as in Deep Q-Networks (DQN) \cite{hernandez2019understanding, van2016deep}. The agent’s behaviour is influenced and adjusted in order to obtain higher rewards, a process that bears some resemblance to the function of neurotransmitters such as dopamine. However, such rewards typically require substantial policy design and are incorporated through a global loss function, rather than directly modulating neuronal activity within the network. As a result, these mechanisms remain distant from true neurotransmitter-inspired processes, as they adjust the network only at a global level and lack the capacity to fine-tune the behaviour of individual neurons in the manner observed in biological systems.

Additionally, neurotransmitter dynamics are inherently \textbf{history-dependent}. Several existing studies relate to this perspective, with the closest line of work being replay-based approaches in continual learning. As new tasks arrive, the network revisits previously learnt information in order to mitigate catastrophic forgetting, a mechanism commonly referred to as experience replay \cite{rolnick2019experience}. However, such approaches are explicitly task-dependent, requiring the identification and revisiting of past tasks or samples. In contrast, neurotransmitters promise modulation relying on an implicit, continuously evolving history state rather than discrete task boundaries. This mechanism bears some resemblance to the state cell of LSTM networks; nevertheless, the underlying principles differ fundamentally. LSTM state cells are primarily designed to regulate gradient flow and decouple long-range dependencies in order to prevent gradient explosion or vanishing. Neurotransmitter-based modulation, by contrast, operates through state-dependent dynamics that continuously influence learning and adaptation without explicitly isolating past knowledge.

Moreover, neurotransmitters coordinate learning across both global and local systems. Previous work has investigated related paradigms, including modular networks \cite{kirsch2018modular} and multi-agent systems \cite{li2024survey}, which decompose complex problems into subtasks and coordinate them at a global level. At the level of local process coordination, Mixture-of-Experts (MoE) architectures have demonstrated substantial effectiveness in practice, and are now widely adopted in large language models, such as DeepSeek \cite{liu2024deepseek}. In addition, parameter masking and plasticity-based algorithms provide an alternative means of achieving selective training by modulating which parameters remain trainable during learning \cite{bai2022parameter}, thereby improving training efficiency and stability. Unlike these architectures, which primarily separate or gate computational pathways, neuromodulatory mechanisms tightly couple global learning signals with local plasticity control, enabling coordinated regulation across multiple hierarchical levels of the system.

\section{Implications}
Applying neurotransmitter-inspired mechanisms in ANNs could carry significant
implications across interdisciplinary research between biology and computer science. 
First, from a technical perspective, neuromodulatory learning has the potential to address fundamental challenges in lifelong learning, including catastrophic forgetting and the management of short- and long-term memory. By enabling selective, state-dependent plasticity, such mechanisms may allow networks to retain essential knowledge while adaptively discarding less relevant information over time.

Second, this paradigm opens a new frontier for dynamic learning, moving beyond static optimisation objectives towards models that can regulate when, where, and how learning occurs. Neurotransmitter-inspired modulation introduces an additional control axis that complements representation learning, enabling more flexible, robust, and context-aware adaptation in complex and evolving environments.

Third, this line of research strengthens the bridge between biological learning systems and computational NNs. By translating well-established principles from neuroscience into computational abstractions, neurotransmitter-inspired models encourage deeper interdisciplinary collaboration and provide a unifying language for integrating insights from neuroscience, machine learning, and cognitive science.

From another perspective, ignoring neuromodulatory principles could limit ANN evolution in the long run. Relying on a single, loss-centric optimisation paradigm runs the risk of reinforcing brittle learning dynamics as models continue to grow and are expected to function in open-ended, lifelong learning environments. Networks are compelled to implicitly encode task importance, stability, and plasticity within parameters or architectures in the absence of an explicit mechanism for state-dependent modulation. This results in inefficient use of capacity, entangled representations, and increased vulnerability to catastrophic forgetting.

Furthermore, learning systems are restricted to static or pre-established adaptation rules in the absence of a neuromodulatory axis, which limits their flexibility in responding to non-stationary environments, changing task priorities, or delayed consequences. This inflexibility could help to explain why continual learning solutions frequently rely more on external memory systems or increasingly intricate architectural expansions than on principled internal regulation of learning dynamics. 

From a broader perspective, overlooking neurotransmitter-inspired modulation risks narrowing the conceptual toolkit of machine learning to representations and objectives alone, while disregarding mechanisms that govern when, where, and how learning should occur. As a result, future systems may achieve impressive performance within fixed training regimes yet remain fundamentally misaligned with the demands of autonomous, adaptive intelligence. The future development of ANNs will be strictly limited to score improvement blindly. Addressing this missing dimension is therefore likely not merely an optional enhancement, but a necessary step toward more robust, scalable, and biologically plausible learning systems.

\section{Open Challenges and direction}

Several open directions and challenges remain for this line of research. First, substantial empirical effort is required to evaluate neurotransmitter-inspired mechanisms across a diverse range of learning problems, and across multiple application domains such as computer vision, audio understanding, and natural language processing. In particular, it is essential to establish whether neuromodulatory mechanisms provide consistent benefits when learning from large-scale, real-world datasets.

Second, further investigation is needed into the behavioural dynamics of artificial neurotransmitters beyond the formulations proposed in this work. This includes a number of fundamental research questions: how many distinct neurotransmitter signals are sufficient; how multiple neurotransmitters interact or co-modulate learning dynamics; and under what conditions such signals should be activated or suppressed during training.

Third, appropriate evaluation metrics must be developed to assess the effectiveness of neuromodulatory learning. Beyond standard accuracy-based measures, additional metrics are needed to capture properties such as plasticity, learning efficiency, such as time to convergence, parameter efficiency, and memory utilisation. Moreover, higher-level qualitative attributes, including stability and flexibility, may also provide valuable insights into neuromodulatory behaviour. Developing unified evaluation protocols that incorporate these factors remains an open challenge.

Finally, interpretability constitutes an important direction for future work. Analysing and visualising the internal representations of neuromodulated neural networks may help determine whether and how neurotransmitter-inspired signals influence learning dynamics. Such analyses are crucial for understanding the functional roles of artificial neurotransmitters and for validating their intended effects.

\section{Conclusion}

We introduce the idea of incorporating neurotransmitter-inspired mechanisms, as found in mammalian brains, into ANNs. We provide an overview of their biological roles, principles, and existing efforts to transfer these concepts into ANNs, and propose general strategies for implementation. Motivated by the challenges in current training paradigms, we advocate a dynamic, biologically inspired learning approach that introduces modulatory signals, such as gain, attention temperature, learning-rate multipliers, or plasticity gates, via a modulator network interacting with a base network (e.g., a policy, value, or predictor model). This design embodies the ``neuromodulated plasticity learning" motif, where eligibility traces are gated by modulatory signals, and can be trained using gradient descent or reinforcement learning. By providing this conceptual blueprint, our perspective aims to enrich learning dynamics and potentially improve multiple dimensions of ANN performance, including stability, adaptability, and lifelong learning.

\sloppy
\bibliographystyle{IEEEtran}
\bibliography{reference}

\end{document}